\pdfoutput=1
\documentclass[10pt,twocolumn,letterpaper]{article}
\usepackage[pagenumbers]{cvpr}
\usepackage{times,graphicx,amsmath,amssymb,booktabs,array}
\usepackage{enumitem}
\usepackage{tikz}
\usetikzlibrary{arrows.meta,positioning,calc}
\definecolor{cvprblue}{rgb}{0.21,0.49,0.74}
\definecolor{fpTeal}{HTML}{14717A}
\definecolor{fpRust}{HTML}{AD5225}
\usepackage[pagebackref,breaklinks,colorlinks,allcolors=cvprblue]{hyperref}
\title{Weight-Space Programming for Vision--Language Model Repair}
\author{Avinash Goutham Aluguvelly\\The University of Texas at Austin\\{\tt\small algo@utexas.edu}}

\begin{document}
\maketitle

\begin{abstract}
In weight-space programming, a language model edits a network's weights directly, without gradients. We ask what a successful repair of this kind relies on. We plant faults in the weights of two vision--language models, SmolVLM-256M and LFM2-VL-450M, and check each edit made by a coding agent against the original weights. Without being told where to look, the agent finds rescaled projector rows and noise in the vision encoder, infers the unknown scale factors, and returns most or all of the rescaled weights to their exact original values and accuracy to near its original level. These repairs depend on a trace of how the faults were computed. The original weights have 16-bit precision, and faults computed in 32-bit arithmetic leave low bits that only the damaged weights carry, a cue that a real failure need not leave. When we round those bits away, each model answers our test questions as before, but the agent, unaided, no longer repairs either model: on LFM2-VL accuracy recovers by about 4 points, against 17 from the unrounded weights. Marking the damaged weights again in their low bits brings the edits back to them. Rounded weights still hold statistical evidence of the rescaling. On SmolVLM the agent uses it once a sentence in the prompt, which we wrote knowing the fault, names such a statistic; on LFM2-VL, whose scale factors are larger, most sessions find the rescaled rows unaided, but none finds the noise. Swapped rows or columns, which keep the set of stored values, are not found unaided on either model within our budgets. We recommend that studies of weight repair report the number format of the weights and include a rounded control.
\end{abstract}

\section{Introduction}
\label{sec:intro}

Neural network weights are usually changed by gradient-based training. \emph{Weight-space programming}~\cite{takehi2026wsp} is a recent alternative in which a language model reads the weights and writes code that edits them. Its results have been judged mainly by the behavior of the edited network. We ask what a successful edit of a vision--language model (VLM) relies on: which evidence the agent finds among hundreds of millions of stored numbers, and what remains of the repair when that evidence is removed.

\begin{figure}[t]
\centering
\input{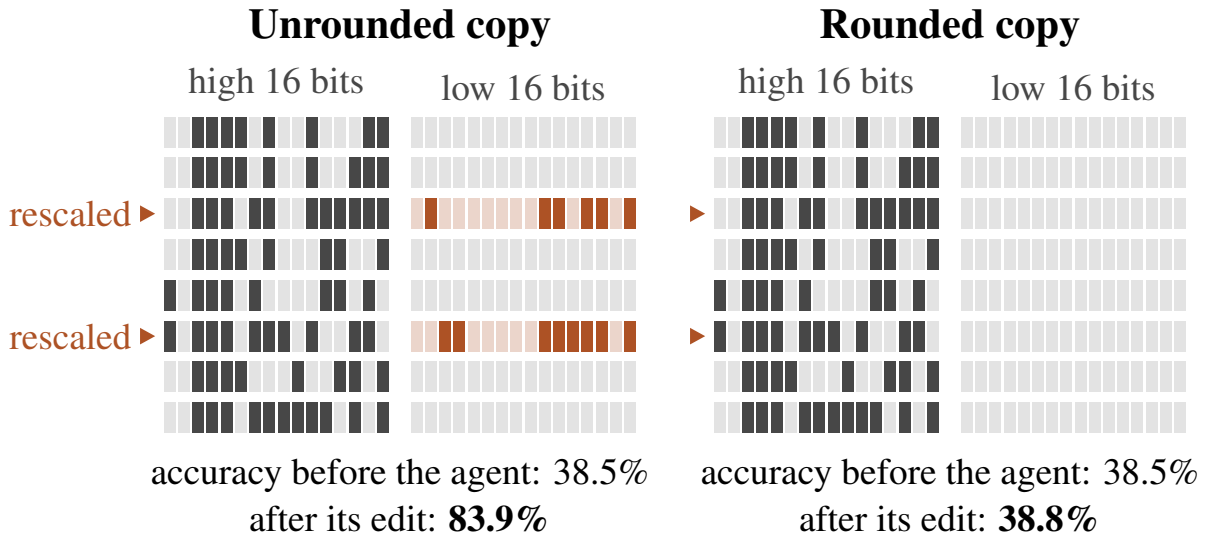}
\caption{A fault computed in float32 leaves low bits that only damaged weights carry. Eight weights from the SmolVLM projector, as 32-bit floats, one cell per bit (dark or orange: 1); the two marked weights belong to rescaled rows. Clean weights take only bfloat16 values, so their low 16 bits are zero. Rounding clears the low bits of the damaged weights and leaves the rows rescaled. The model gives the same answers to all test questions in both copies, but the agent repairs only the unrounded one (mean of two sessions on each of checkpoints A--C; clean accuracy 83.3\%).}
\label{fig:lowbits}
\end{figure}

We study this question using model repair, because a repair can be checked against ground truth. We use two VLMs: SmolVLM-256M~\cite{marafioti2025smolvlm}, adapted to spatial questions about synthetic scenes, and LFM2-VL-450M~\cite{liquidlfm450,amini2025lfm2}, adapted to spatial questions about GQA photographs~\cite{hudson2019gqa}. In each we plant faults in the vision encoder, in the projector that connects vision to language, and in the language model. In each \emph{session}, a coding agent receives a damaged checkpoint, a few labeled examples, and a limited number of accuracy scores. It may read any weight, must return a program that edits weights, and may not use gradients. The clean weights, the fault locations, and the test images are withheld. Because we planted the faults, we can check each edit weight by weight: which weights it changed and whether it restored them. Most SmolVLM conditions are repeated over three to six checkpoints. On LFM2-VL the central comparison is repeated on two further checkpoints, and every other condition is a single session.

With the damaged weights as the faults left them, the agent repairs both models. On SmolVLM, six sessions on three checkpoints change only damaged weights and raise spatial accuracy from 38.5\% to 83.9\% (clean: 83.3\%). On LFM2-VL, eight sessions on two checkpoints do the same and raise accuracy from 55.5\% to 72.5\% (clean: 75.0\%). The agent recovers the unknown scale factors from the weights, and in thirteen of these fourteen sessions it returns every rescaled weight to its exact clean value.

Our main finding is that this repair depends on the precision of the stored numbers. The public weights of both models are stored in bfloat16 (BF16), a 16-bit format, and so is every weight of our clean checkpoints. The checkpoints are held, edited, and run as 32-bit floats, in which the low sixteen bits of every clean weight are zero. A fault computed in 32-bit arithmetic fills those bits, so the damaged weights become the only ones that use them (Fig.~\ref{fig:lowbits}). The agent is not told this; the sessions find it and use it both to locate the damage and to recover each factor. A fault planted in this way thus hands the agent a cue that a failure of a checkpoint stored in 16 bits need not leave.

When we round the damaged weights back to 16 bits, the damage remains and each model gives the same answer to every test question, but the agent no longer repairs either model. On SmolVLM five of six sessions return no edit. On LFM2-VL, whose scale factors are larger, four of six sessions still find the rescaled rows by their magnitudes, but none finds the noise, which only the low bits revealed, and accuracy rises by 3.6 points, against 17.1 from the unrounded copies. When we set the low bits of the damaged weights again, by the smallest step a 32-bit float allows, the edits return to those weights on both models.

Rounded weights still hold evidence of the scaling fault. Multiplying by a constant and rounding skips some values: multiplying the whole numbers by 1.4 and rounding never gives 2, 5, or 9. For the same reason, a rescaled and rounded row usually lacks some of the 128 values that its seven stored fraction bits can take (Fig.~\ref{fig:hist}), and its norm changes. On SmolVLM the sessions do not use this evidence until a one-sentence hint, which we wrote knowing the fault, names such a statistic without a location. The hint directs the agent's edits to the rescaled rows and, on six checkpoints, gains 19.0 points more than a placebo sentence. Swapped rows or columns leave no such evidence, because they keep the set of stored values, and the agent did not find them unaided on either model within our budgets.

Checking edits against the clean weights separates outcomes that accuracy does not. A fixed script recovers most or all of the agent's gain on unrounded weights. Gradient fitting recovers more accuracy than the hinted agent on rounded weights but, as reported for small networks~\cite{nusrat2026clinical}, changes nearly every weight it may edit and restores almost none. Programs that the agent writes from unrounded weights restore 98.6--100\% of the rescaled weights of unseen checkpoints of the same model without feedback, whereas a program written from rounded LFM2-VL weights repairs no unseen combined fault.

\section{Related Work}
\label{sec:related}

\paragraph{Weight-space programming.}
Takehi et al.~\cite{takehi2026wsp} had language models write policies for MinAtar games and Go into zero-initialized networks, edit two matrices of a Qwen text model to reduce reasoning loops, and identify trained behaviors from weights. Follow-up studies apply the approach to robot policies~\cite{zhang2026robots}, medical predictors~\cite{lindholz2026medical}, online updates~\cite{li2026weightloop}, and policies checked by a model checker~\cite{gross2026verification}, and measure variation across sessions~\cite{hu2026exploring}. Ahmed~\cite{ahmed2026audit} finds that a language model's readings of small networks rest largely on local arithmetic and memorized priors. Closest to our setting, Nusrat and Nusrat~\cite{nusrat2026clinical} plant flaws in small clinical networks through the training data; a language model repairs most of them by changing a few parameters, whereas fine-tuning changes nearly all. We plant faults in VLM weights, so that each edited weight can be compared with its clean value, and remove and reintroduce numerical evidence to test what the repair relies on. Model editing methods instead derive edits from data, activations, or gradients~\cite{usman2021nnrepair,ilharco2023arithmetic,meng2022rome,santurkar2021rewriting}.

\paragraph{Forensic traces.}
The pattern that remains in rounded weights (Sec.~\ref{sec:histogram}) is known from image forensics. A JPEG that is quantized twice with different steps shows periodic empty and doubled bins in its coefficient histograms~\cite{popescu2004forensics}, from which the first quantization can be estimated~\cite{lukas2003double}. A weight row that is rescaled and rounded to BF16~\cite{kalamkar2019bfloat16} is requantized likewise.

\section{Task and Audit Protocol}
\label{sec:setup}

\begin{figure*}[t]
\centering
\input{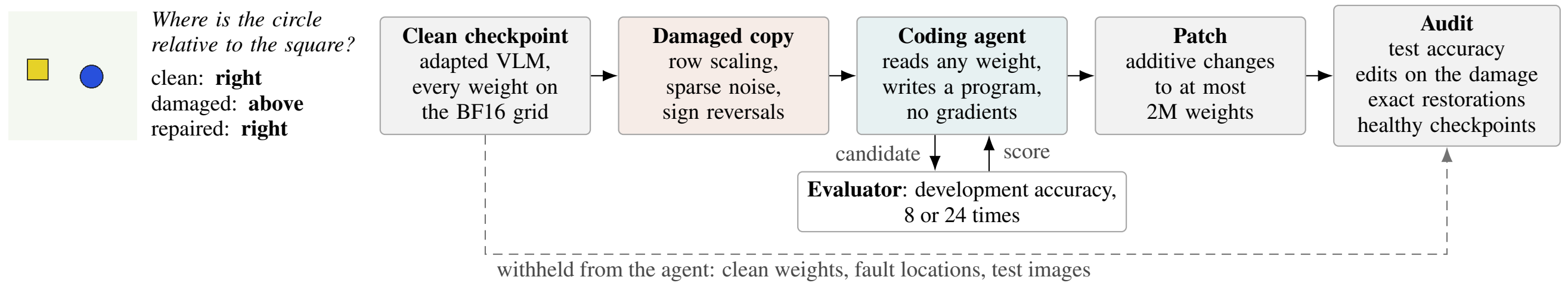}
\caption{The repair task, shown for SmolVLM with a test image from checkpoint~A. The agent sees only the damaged checkpoint, a few labeled examples, and scores for candidate patches; its final patch is audited against the clean weights.}
\label{fig:protocol}
\end{figure*}

\paragraph{Models and tasks.}
Each experiment damages a clean checkpoint, gives a copy to the agent, and audits the returned patch against the clean weights (Fig.~\ref{fig:protocol}). We run this protocol on two testbeds (Tab.~\ref{tab:testbeds}). In the first, we generate the images ourselves: each $256\times256$ image shows a circle and a square of different colors. The main question asks where the circle is relative to the square, and the answer is \emph{left}, \emph{right}, \emph{above}, or \emph{below}. The public SmolVLM-256M-Instruct checkpoint answers about 16\% of these questions correctly, so we adapt it by training a $128\times128$ block of its projector matrix on 32 spatial and 16 color examples.

The second testbed uses photographs. Its main questions are GQA questions that mention exactly one of the same four relations and are answered \emph{yes} or \emph{no}. The test set has 48 questions with each answer, so a constant answer scores 50\%. The public LFM2-VL-450M checkpoint answers 70.8\% of our spatial test questions, and we adapt it by training a $128\times128$ block of the second matrix of its projector on 64 photographs.

In both testbeds, further question sets check for side effects: colors, the spatial questions in other words, and, on photographs, quantities. Repeating the adaptation with independent seeds and, except for L3 and L4, new images gives further \emph{clean checkpoints}, which we round to BF16 (Tab.~\ref{tab:testbeds}); L1 had served in an earlier pilot with other test photographs (Supp.~S21). Answers are generated by greedy decoding and count as correct when they match the expected word after normalization (Supp.~S3; the supplement is an ancillary file of this preprint).

\begin{table}[t]
\centering
\footnotesize
\setlength{\tabcolsep}{3pt}
\begin{tabular}{@{}lll@{}}
\toprule
 & SmolVLM-256M & LFM2-VL-450M\\
\midrule
Projector matrix & $576\times12{,}288$ & $1{,}024\times2{,}560$\\
Scale factors & 0.50--0.74, 1.35--2.01 & 0.6, 0.725, 3.1, 5.15\\
Checkpoints, Sec.~\ref{sec:trace} & A--C & L1, L3, L4\\
\quad Secs.~\ref{sec:histogram}, \ref{sec:fitting} & A--C, T4--T9, R1--R3 & L1\\
\quad Sec.~\ref{sec:program} & T1--T3 & L2--L4\\
Clean accuracy & 69--97\% & 75.0--79.2\%\\
Spatial test set & 24--64 questions & 96 questions\\
Session budget & 24 scores, 30 min & 8 scores, 15 min\\
\bottomrule
\end{tabular}
\caption{The two testbeds. Scale factors are those of the combined fault in Secs.~\ref{sec:trace}--\ref{sec:fitting}; accuracies and test-set sizes are ranges over checkpoints.}
\label{tab:testbeds}
\end{table}

\paragraph{Faults.}
We plant four kinds of fault. The first three have a strength $s$ and lie in different parts of the model:
\begin{itemize}[leftmargin=1.1em,itemsep=1pt,topsep=3pt,parsep=0pt]
\item \textbf{Scaling} (projector). A quarter of the rows of the projector matrix are each multiplied by a factor $g$ (Tab.~\ref{tab:testbeds}). On SmolVLM every row has its own factor, $g=e^{su}$ or $e^{-su}$ with $u$ uniform on $[0.3,0.7]$; on LFM2-VL four factors $g=1+s(g_0-1)$, with $g_0\in\{0.2,0.45,5.2,9.3\}$, apply to 64 rows each.
\item \textbf{Noise} (vision encoder). Gaussian values with standard deviation $6s$ times that of the tensor are added to 1\% of the weights in one MLP matrix.
\item \textbf{Sign reversal} (language model). Rows of one MLP matrix are negated.
\item \textbf{Swaps} (projector). Pairs of rows or pairs of columns of the projector matrix exchange places.
\end{itemize}
The \emph{combined} fault applies the first three and touches 0.7\% of SmolVLM's weights and 0.2\% of LFM2-VL's. Strengths are set on separate calibration checkpoints; the combined faults in Secs.~\ref{sec:trace}--\ref{sec:fitting} use $s=1$ on SmolVLM and $s=\tfrac12$ on LFM2-VL.

\paragraph{BF16 grid and rounding.}
BF16 keeps the sign, the eight exponent bits, and the top seven fraction bits of a float32. Written back as float32, a BF16 value has sixteen zero low bits (Fig.~\ref{fig:lowbits}). We call the float32 values that BF16 represents exactly the \emph{BF16 grid}; an \emph{off-grid} value has nonzero low bits. Every checkpoint is a float32 file: the agent reads float32 values, and patches are added and the model is run in float32. Scaling and noise faults computed in float32 produce off-grid values, while a sign reversal stays on the grid. From each damaged checkpoint we make two copies: the \emph{unrounded} copy keeps the values the faults produced, and the \emph{rounded} copy rounds each value to the nearest grid point. Rounding moves each weight by at most 0.4\% of its magnitude and leaves the damage in place. On checkpoints A--C, L1, L3, and L4, at most seventeen of the 0.8 to 1.8 million damaged weights return to their clean values, and the two copies give the same answer to every test question. The agent's starting feedback is nearly the same as well: the development accuracies of the two copies are identical, and their repair losses differ by at most 0.12\% (Supp.~S24).

\begin{table*}[t]
\centering
\footnotesize
\setlength{\tabcolsep}{4.5pt}
\begin{tabular}{@{}lrrcrrrcr@{}}
\toprule
 & \multicolumn{4}{c}{SmolVLM-256M} & \multicolumn{4}{c}{LFM2-VL-450M}\\
\cmidrule(lr){2-5}\cmidrule(l){6-9}
Fault, version of the weights, and editor & Start & After & Localized & Restored & Start & After & Localized & Restored\\
\midrule
\multicolumn{9}{@{}l}{\emph{Combined fault} (Sec.~\ref{sec:trace})}\\
\quad Unrounded, agent & 38.54 & 83.85 & 6/6 & 98.4\% & 55.47 & 72.53 & 8/8 & 84.4\%\\
\quad Unrounded, grid-fit script & 38.54 & 79.17 & 3/3 & 78.8\% & 55.47 & 73.70 & 4/4 & 0.0\%\\
\quad Rounded, agent & 38.54 & 38.80 & 0/6 & 0.0\% & 53.82$^\ast$ & 57.47 & 4/6 & 24.7\%\\
\quad Rounded, marks on damage, agent & 46.88 & 72.92 & 3/3 & 0.3\% & 47.92 & 76.04 & 1/1 & 0.0\%\\
\quad Rounded, marks on decoys, agent & 46.88 & 45.83 & 0/3 & 0.0\% & 47.92 & 47.92$^\ddagger$ & 0/1 & 0.0\%\\
\quad Rounded, random low bits on damage, agent & 38.54 & 79.69 & 6/6 & 0.0\% & -- & -- & -- & --\\
\quad Rounded, random low bits on decoys, agent & 38.54 & 56.51 & 4/6 & 0.1\% & -- & -- & -- & --\\
\addlinespace
\multicolumn{9}{@{}l}{\emph{Swaps in the projector matrix, agent} (Sec.~\ref{sec:swap})}\\
\quad Row swaps, no hint & 23.44 & 28.12 & 0/1 & 0.0\% & 66.67 & 68.75 & 0/1 & 0.0\%\\
\quad Row swaps, fault named & 23.44 & 23.44$^\dagger$ & 0/1 & 0.0\% & 66.67 & 65.62 & 0/1 & 0.0\%\\
\quad Column swaps, no hint & 57.81 & 62.50 & 0/1 & 0.0\% & 69.79 & 64.58 & 0/1 & 0.0\%\\
\quad Column swaps, fault named & 57.81 & 76.56 & 1/1 & 99.7\% & 69.79 & 67.71 & 0/1 & 0.0\%\\
\quad Row and column swaps, no hint & 43.75 & 43.75$^\dagger$ & 0/1 & 0.0\% & 72.92 & 77.08 & 0/1 & 0.0\%\\
\quad Row and column swaps, fault named & 43.75 & 43.75$^\dagger$ & 0/1 & 0.0\% & 72.92 & 72.92 & 0/1 & 0.0\%\\
\bottomrule
\end{tabular}
\caption{Repairs by fault, version of the weights, and editor. \emph{Start} and \emph{After} are spatial test accuracy (\%) before and after the edit, averaged over sessions and then checkpoints. \emph{Localized} counts runs whose patch puts at least 95\% of its changes on damaged weights. \emph{Restored} is the share of all damaged weights returned to their exact clean values. Marks, random low bits, and decoys are defined in Sec.~\ref{sec:marks}. SmolVLM: A--C (64 test questions each; clean accuracy 83.33\%), with 32 other questions and smaller budgets in the mark rows (87.50\%), and, for swaps, T6 (76.56\%). LFM2-VL (96 test questions): L3 and L4 in the first three rows (75.00\%; four fault draws, two sessions each) and L1 below (79.17\%; one session per row). $^\ast$Six sessions on three draws; two received no scores (Sec.~\ref{sec:residue}). No valid selected patch (Sec.~\ref{sec:setup}): $^\dagger$the session submitted a patch other than the one its scores select; $^\ddagger$a file written by the agent disabled the evaluation tools.}
\label{tab:views}
\end{table*}

\paragraph{Agent interface.}
We use GPT-6 Astra (\texttt{gpt-6-astra}), through the Codex CLI, as the coding agent. Each run on one checkpoint is a \emph{session}, held in an isolated container at high reasoning effort (low effort for the SmolVLM sessions of Secs.~\ref{sec:trace} and~\ref{sec:program}). Unless noted, a session has:
\begin{itemize}[leftmargin=1.1em,itemsep=1pt,topsep=3pt,parsep=0pt]
\item \textbf{Inputs.} Read-only access to the damaged checkpoint, the model configuration, and sixteen labeled repair examples. The prompt states that no fault location or reference checkpoint is supplied and that an edit may be unnecessary. The public checkpoint is not available.
\item \textbf{Output.} A program that produces a \emph{patch}: additive changes to at most two million weights (SmolVLM's combined fault alters 1.8 million).
\item \textbf{Feedback.} An evaluator applies a candidate patch and returns only aggregate scores: accuracy on 16 spatial development images, a second accuracy (16 color images on SmolVLM, 16 further spatial images on LFM2-VL), and the repair loss. Budgets are in Tab.~\ref{tab:testbeds}.
\item \textbf{Selection.} The first candidate with the best spatial development accuracy is final (ties: second accuracy, then repair loss); the empty patch is always a candidate.
\end{itemize}

\paragraph{Audit.}
When a session ends we freeze its program, confirm offline that it reproduces the patch byte for byte, apply the patch, reload the model, and score it on test images. With the clean weights we then count the edited weights that were damaged and the damaged weights whose new value equals the clean value bit for bit (\emph{exact restorations}). Clean checkpoints given as inputs are called \emph{healthy}. We report accuracy changes in percentage points, averaging sessions within a checkpoint before averaging checkpoints. Apart from L3 and L4 in Sec.~\ref{sec:residue}, each LFM2-VL entry is one session on one checkpoint; one test answer is 1.04 points, and Supp.~S27 gives the answers gained and lost by each edit. A session without a valid selected patch, because it failed or submitted a patch other than the one its scores select, leaves the checkpoint unchanged; we count it as no edit and mark it in the tables. Sessions that received no scores through a failure of our evaluation service are excluded.

\section{The Role of the Low Bits in Repair}
\label{sec:trace}

We vary the low bits of the agent's copy of the combined fault (Tab.~\ref{tab:views}, which also lists the swaps of Sec.~\ref{sec:swap}).

\subsection{Unrounded versus rounded faults}
\label{sec:residue}

\paragraph{Unrounded.}
On SmolVLM, each unrounded copy of checkpoints A--C goes to two fresh sessions, and all six follow the same procedure. They notice that a few tensors hold values with nonzero low bits. For each off-grid projector row they estimate a scale factor, divide it out, and round the row back to the grid; they shrink or zero the sparse off-grid values in the vision encoder. Every patch changes 1.79 million weights, all of them damaged and 99.74\% of all damaged weights, and returns all 1,769,472 rescaled weights to their exact clean values. No noise value is restored exactly, and the 4,608 sign-reversed weights, which stay on the grid, are left unchanged. Spatial accuracy rises from 38.54\% to 83.85\%; the clean checkpoints score 83.33\%.

On LFM2-VL, L3 and L4 each receive two independent draws of the combined fault, with the same four factors on new rows and positions, and each draw goes to two sessions per copy. These eight sessions on unrounded copies and a ninth on L1 proceed as on SmolVLM and change only damaged weights. They identify four groups of 64 projector rows scaled by 0.6, 0.725, 3.1, and 5.15, the planted factors, and divide them out. Seven of the eight sessions on L3 and L4 return all 655,360 rescaled weights to their exact clean values; the eighth and the L1 session, which do not round the corrected rows to the grid, return 94\% and 84\%. Eight of the nine also shrink or zero the off-grid entries of the vision matrix; they reach 70 to 73 correct answers of 96 on L3 and L4, where the clean checkpoints give 72, and 75 on L1, one below the clean checkpoint. The ninth, on L4, corrects the rescaled rows alone and gains no answer. Over the eight sessions on L3 and L4 the mean gain is 17.06 points; two kinds of random control edits of the same size lose 0.13 and 1.56. No session changes the sign-reversed rows.

\paragraph{How a factor is identified.}
The low bits fix a row's factor only up to a power of two: if $w$ is a grid value, so is $w/2$, so a row scaled by $g$ also returns to the grid when it is divided by $2g$ or $g/2$. The sessions, and the programs of Sec.~\ref{sec:program}, choose the power that brings the row's norm closest to those of the unaffected rows. Exact restoration thus also assumes that a matrix's rows have similar norms.

\paragraph{Rounded.}
With the rounded SmolVLM copies, five of the six sessions test candidates, find none that helps, and return the empty patch. The sixth edits healthy weights, gains one spatial answer, and loses fifteen color answers. Three further sessions at high effort gain nothing (Supp.~S8).

On LFM2-VL most sessions find the scaling fault, but none repairs the model. Two of the eight sessions on rounded copies of L3 and L4, both on one fault draw, received no scores because our evaluation service rejected their calls, and we exclude them. The other six gain 3.65 points (controls: 1.04), against 17.53 for the unrounded sessions on the same three draws; on each draw the unrounded sessions gain more, by 10 to 16 answers on average. Four of the six edit only rescaled rows, which stand out by their magnitudes. Two of these, on the same draw, also identify the four factors: the candidate values for which dividing a row, rounding it to the grid, and multiplying back reproduces the stored weights. Their edits return 87\% of the rescaled weights to their exact clean values. A fifth session reports no defect and multiplies the input embeddings of the \emph{yes} tokens by 1.075, which gains twelve answers. The L1 session does the same with 1.15: the model, which answered \emph{no} to 94 of the 96 questions, then answers \emph{yes} 62 times, so the gain of 8.33 points reflects a shift in the answer bias. The sixth session edits healthy tensors.

None of these seven sessions changes a noise weight, and the noise accounts for the difference in accuracy. On the draw where two rounded sessions restored the rescaled rows, the unrounded session that did the same and left the noise in place also stays at 53 correct answers, while the session that zeroed the noise reaches 73. With the noise fault alone, a session on L1 gains 12.50 points from the unrounded copy, by changing only off-grid entries, and 2.08 from the rounded one. On LFM2-VL the low bits thus matter for the sparse noise, for which Sec.~\ref{sec:histogram} offers no row statistic, while the large scale factors can be found without them.

\paragraph{Fixed script and other question sets.}
A fixed \emph{grid-fit} script that we wrote after seeing these sessions, which searches for a divisor that returns each off-grid row to the grid, does most of what the agent does (Tab.~\ref{tab:views}). On SmolVLM, with eight or nine development scores, it changes as many weights as the agent, all of them damaged, and gains 40.63 of the agent's 45.31 points. A simpler script for LFM2-VL, without feedback, gains 26.04 of the agent's 30.21 points on L1 and 18.23 points on L3 and L4, where the sessions gain 17.06. Neither script edits a rounded or a healthy checkpoint. On A--C and L1 the agent's repair from unrounded weights also returns the color, reworded, and quantity questions to their clean accuracy (Supp.~S22, S23). Four sessions on the healthy L3 and L4 lose no answer on any set (Supp.~S25).

\subsection{Marking the damaged weights}
\label{sec:marks}

Rounding also changes the weight values slightly. To isolate the effect of the low bits, we start from the rounded checkpoints and move each damaged weight one float32 step away from zero. No mark is larger than $1.2\times10^{-7}$, and the marks leave the test answers and the development accuracies unchanged. As a control, the same number of marks is placed on healthy weights of the same tensors (\emph{decoys}). On SmolVLM each version gets one session per checkpoint, with eight scores, 20 minutes, eight development images per set, no labeled examples, and 32 new test images.

All four sessions that see marks on the damage change only damaged weights. On A and L1 the agent edits all three damaged tensors; on B it reverses only the marked sign rows, which restores all 4,608 reversed weights exactly and costs three answers. With marks on decoys or with no marks, no SmolVLM patch touches a damaged weight; the decoy session on L1 could not submit its patch.

The marks carry location only, whereas the low bits that a fault leaves also indicate the correction. When we repeat the SmolVLM experiment of Sec.~\ref{sec:residue} after replacing the low bits of each damaged weight with random bits that round to the same grid point, all six patches lie entirely on damaged weights. They gain 41.15 points, against 45.31 with the original low bits, but restore at most two weights each. Location therefore accounts for most of the gain in accuracy, and the values of the low bits for the exact restoration. With random low bits on decoys instead, four of six sessions find the real rescaled rows, from unusual row norms and from gaps in the fraction bits of the \emph{on-grid} values.

\section{Repair from Rounded Weights}
\label{sec:histogram}

After rounding, the low bits carry no information. We ask what else marks the damage and whether the agent uses it.

\subsection{What rounded weights retain}

\begin{figure}[t]
\centering
\input{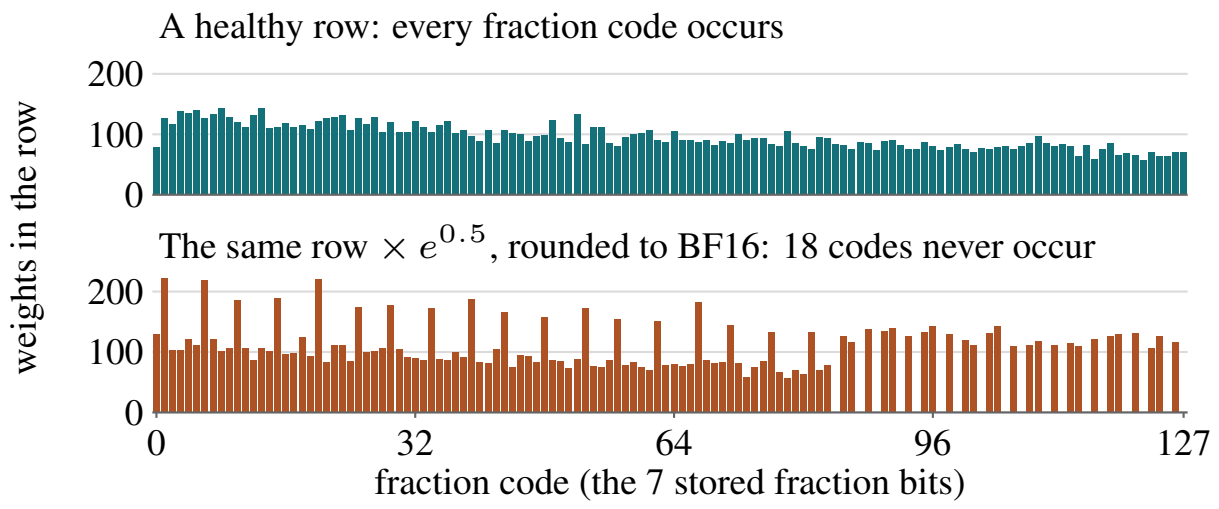}
\caption{Fraction codes in one row of the SmolVLM projector (12,288 weights), before and after the row is multiplied by $e^{0.5}$ and rounded to BF16. The rescaled row leaves 18 codes unused.}
\label{fig:hist}
\end{figure}

A normal BF16 value has magnitude $(1+c/128)\,2^{e}$, where the \emph{fraction code} $c\in\{0,\dots,127\}$ holds the seven stored fraction bits. A power-of-two factor leaves the codes unchanged, so consider a factor $g$ with $1<g<2$. Multiplying by $g$ and rounding spreads apart the codes whose product stays below the next power of two, so that some codes are never produced, and merges those whose product crosses it (Fig.~\ref{fig:hist}). If the row uses every code, about $128\,(2-g)(g-1)/g$ codes stay empty: at most 22, for $g=\sqrt2$, and 18 in Fig.~\ref{fig:hist}. No row of the public SmolVLM projector has an empty bin.

Both statistics detect rescaled rows reliably. On rounded T4--T9, ranking projector rows by the distance between their fraction histogram and the median row histogram separates the rescaled rows with an area under the ROC curve (AUC) of 0.992--1.000, and ranking by row norm gives the same range. On L1 both exceed 0.999. For sign-reversed rows both are at chance, and neither applies to the sparse noise, which touches every row.

\subsection{Effect of a prompt hint}
\label{sec:hint}

Given rounded SmolVLM weights and neither a hint nor a location, agents used these statistics only in the sessions with random low bits on decoys (Sec.~\ref{sec:marks}). In a pilot on A--C we added one sentence to the prompt: \emph{``General diagnostic suggestion: consider whether histograms of the BF16 fraction bits, with the sign and exponent removed, differ across rows in weight matrices.''} Accuracy then rose by 26.56 points, with every edit on rescaled rows. Because we wrote that sentence after studying A--C, we fixed the design and ran it on six new checkpoints, T4--T9. Each rounded, damaged checkpoint gets two sessions under each of three prompts: no hint, the sentence above, and a placebo that names the sign bit in place of the fraction bits. Neither names a tensor, row, or fault. The placebo controls for adding a sentence; the choice of statistic was ours, made knowing the fault.

\begin{table}[t]
\centering
\footnotesize
\setlength{\tabcolsep}{5pt}
\begin{tabular}{@{}lrrr@{}}
\toprule
 & \multicolumn{2}{c}{SmolVLM} & LFM2-VL\\
\cmidrule(lr){2-3}\cmidrule(l){4-4}
 & T4--T9 & R1--R3 & L1\\
\midrule
Start accuracy (\%) & 42.19 & 46.35 & 47.92\\
\midrule
\multicolumn{4}{@{}l}{\emph{Agent, by prompt}}\\
\quad No hint & 0.00\rlap{$^\dagger$} & -- & 8.33\\
\quad Placebo hint (sign bits) & 0.65 & 3.65 & 0.00\\
\quad Fraction hint & 19.66 & 14.06 & 5.21\\
\quad Row-norm hint & -- & 11.20 & 4.17\\
\quad No hint, rows given & -- & 11.46\rlap{$^\ddagger$} & 3.12\\
\multicolumn{4}{@{}l}{\emph{Fixed rules on the projector matrix}}\\
\quad Histogram rule & 4.17 & -- & 3.12\\
\quad Row-norm rule with scores & -- & 10.42 & 0.00\\
\multicolumn{4}{@{}l}{\emph{Gradient fits of the projector matrix}}\\
\quad Adam on the given rows & -- & 36.46 & 4.17\\
\quad LoRA on the matrix & 25.26 & -- & 7.29\\
\quad Adam on the matrix & 36.46 & -- & 13.54\\
\bottomrule
\end{tabular}
\caption{Methods applied to the rounded combined fault: gain in spatial test accuracy, in points. \emph{With scores}: correction strength chosen on development scores. The \emph{given rows} are the projector rows with the most unusual norms, as many as were rescaled, chosen without fault labels. Sessions without a valid patch count as no edit: $^\dagger$one of twelve; $^\ddagger$one of three, one session per checkpoint (17.19 without it). The two entries of 36.46 come from separate fits.}
\label{tab:methods}
\end{table}

On T4--T9 the fraction hint gains 19.66 points and the placebo 0.65 (Tab.~\ref{tab:methods}). The difference of 19.01 points is positive on all six checkpoints, with a two-sided 95\% $t$ interval, paired over checkpoints, of 8.97 to 29.05 and an exact sign-flip $p$-value of $2/64$. Eleven of the twelve hinted sessions submit an edit, each of 1.74 to 1.77 million weights, all in rescaled rows. No other edit touches a damaged weight.

On three further checkpoints, R1--R3 (Supp.~S20), we added a sentence that names root-mean-square (RMS) magnitudes of rows in place of fraction bits. The fraction hint gains 14.06 points, the row-norm hint 11.20, and the placebo 3.65. All six edits under the fraction hint lie on rescaled rows, as do at least 98\% of the changes in three of the five valid edits under the row-norm hint, which gains 15.62 points over its valid sessions.

On L1 each of the three sentences gets one session. With the fraction hint and with the row-norm hint, the agent rescales 128 of the 256 rescaled projector rows and no other weight; with the placebo it edits healthy weights. The model still answers \emph{no} to about 90 of the 96 questions, and the spatial gains are 5.21, 4.17, and 0.00 points, below the 8.33 of the unhinted edit of Sec.~\ref{sec:residue}; unlike that edit, the two hinted ones also raise color accuracy, from 7 of 32 to 16 and 14 (Supp.~S22). Fixed rules without scores that rescale the rows flagged by either statistic do no better under the combined fault. Each gains 3.12 points on L1, and they lose 5.47 and 6.77 on L3 and L4, although on L1 they recover 10.42 to 17.71 points from scaling alone (Supp.~S19, S25).

On LFM2-VL the hint is often unnecessary: with factors up to 5.15, five unhinted sessions edited rescaled rows of rounded weights (Secs.~\ref{sec:residue} and~\ref{sec:program}). On SmolVLM, where the factors lie between about $\tfrac12$ and 2, no session without a hint, low bits on decoys, or given rows did. The hint also increases edits to healthy checkpoints: on healthy T4--T9, hinted agents edit all six and unhinted agents two, with mean accuracy changes of $-1.30$ and $-2.34$ points.

\subsection{Faults that preserve stored magnitudes}
\label{sec:swap}

Here we change the fault, so that neither the low bits nor the fraction histogram marks it. A swap moves weights to other coordinates and leaves the multiset of stored values unchanged: every weight stays on the grid, a column swap leaves each row's histogram and norm unchanged, and a row swap only moves them to another row. We plant swaps in the projector matrix of the clean T6 and L1 checkpoints: on T6 they exchange a quarter of the rows, a quarter of the columns, or an eighth of each, and on L1 half as many. On T6 we also negate a quarter of the rows, which keeps every magnitude; that session leaves accuracy at 25.00\%. Each condition gets one session (Tab.~\ref{tab:views}, bottom); the T6 sessions have eight scores and six minutes, a fifth of the default time.

Without a hint and within these budgets, the agent did not find these faults on either model. The sessions find no off-grid values and edit other tensors, such as position embeddings or the output head; none restores a damaged weight.

A hint that names the kind of fault changes the outcome for columns on SmolVLM. We repeat the swap sessions with one added sentence, such as ``Columns were swapped in a weight matrix,'' which names no tensor (Supp.~S2). With the column hint the agent finds the SmolVLM projector, observes that its 12,288 columns form sixteen blocks of 768 features that normally resemble one another, undoes 1,531 of the 1,536 swaps, and returns accuracy to the clean level. With the row hint the agent does not find the swapped rows, and with both hints it names 767 of the 768 column swaps and no row swap. On LFM2-VL the hints do not help: each hinted session selects a matrix of the vision encoder or the language model and restores nothing.

A sign reversal also preserves magnitudes. Alone, it does not lower the accuracy of L1; told the kind of fault, a session finds all 40 reversed rows from their covariance with neighboring layers but zeroes them.

\section{Fixed Rules and Gradient Fitting}
\label{sec:fitting}

We apply the same audit to simpler repairs of the rounded combined fault (Tab.~\ref{tab:methods}): fixed rules using the statistics of Sec.~\ref{sec:histogram}, and fine-tuning, both told which matrix to edit.

\paragraph{Fixed rules.}
The \emph{histogram rule} ranks the rows of the projector matrix by fraction histogram, takes the known fraction of damaged rows from the top, estimates each row's factor, and divides the row by that estimate raised to a power. The power is 1 without feedback; with as many development scores as the matching agent session used, it is chosen among values from 0 to 1.5. On T4--T9 the rule gains 4.17 and 8.20 points, and the hinted agent is ahead of it on five of six checkpoints. Both edit rescaled rows, with 99.5\% and 100\% of their changes on damaged weights, so the agent's advantage lies in the corrections it computes.

\paragraph{Gradient fits.}
On T4--T9 we train the projector matrix with Adam~\cite{kingma2015adam}, directly and through rank-8 LoRA~\cite{hu2022lora}, using the agent's repair examples, development images, and selection rule. We recalibrated the learning rates on a separate checkpoint after a first run at an untuned rate had been scored on T4--T9 (Supp.~S16). Fitting recovers more accuracy: Adam gains 36.46 points and LoRA 25.26, against 19.66 for the hinted agent. The fits edit the whole matrix, so on R1--R3 we restrict Adam, a row-norm rule, and an unhinted agent to the same 144 rows with the most unusual norms. Adam still gains more than three times as much as the rule or the agent, so the advantage of fitting does not depend on editing more weights. On L1 only Adam on the whole matrix exceeds the unhinted agent.

The edits differ in which weights they change. The fits change essentially every weight they may edit, 7.08 million on SmolVLM and 2.62 million on LFM2-VL when the whole matrix is trained, of which a quarter are damaged, and they return at most one damaged weight to its clean value. Exact restoration is a strict test for a fit. On T4--T9 we therefore also measure how much of the distance between the damaged and the clean projector matrix an edit removes: 84\% on average for the hinted agent, six of whose eleven edits leave under 0.4\% of it, 61 to 78\% for the histogram rule, and none for Adam and LoRA (Supp.~S26). On the healthy versions of T4--T9, Adam edits three of the six and raises mean accuracy by 4.69 points. Fitting thus compensates for the damage without identifying it, and it is the better choice when accuracy is the only goal.

\section{Transfer of Agent-Written Programs}
\label{sec:program}

\paragraph{Design.}
A session's final program may rely on row indices and scale factors saved from its own checkpoint, so we separate writing a program from applying it. For each model, programs are written on one checkpoint and tested on others of the same model that are adapted and damaged with independent seeds, with faults in the same tensors: T1--T3 for SmolVLM, and L2--L4 for LFM2-VL, where the programs are written on L1. On the writing checkpoint, one session sees the unrounded combined fault and one sees the rounded version, and the prompt states that the program will be run on other checkpoints (Supp.~S18, S19). The SmolVLM writing sessions have eight scores, 20 minutes, and no labeled examples, and on T1--T3 the noise and combined faults have $s=\tfrac12$, with scale factors between 0.70 and 1.42. The selected programs are P1 and P2 on SmolVLM and P3 and P4 on LFM2-VL. We freeze the code, verify that it contains no coordinates, masks, or fitted constants from its writing checkpoint, and run it on every test case (Tab.~\ref{tab:transfer}) in an offline container, with no scores, labels, or model.

\begin{table}[t]
\centering
\footnotesize
\setlength{\tabcolsep}{3.3pt}
\begin{tabular}{@{}lrrrrrr@{}}
\toprule
 & \multicolumn{3}{c}{SmolVLM, T1--T3} & \multicolumn{3}{c}{LFM2-VL}\\
\cmidrule(lr){2-4}\cmidrule(l){5-7}
Input & Start & P1 & P2 & Start & P3 & P4\\
\midrule
Scaling & 69.44 & $+20.83$ & $+1.39$ & 64.58 & $+11.46$ & $+12.50$\\
Noise & 83.33 & $+9.72$ & $-6.94$ & 65.62 & $+11.46$ & 0.00\\
Sign reversal & 93.06 & 0.00 & $-5.56$ & 77.08 & 0.00 & 0.00\\
Combined & 70.83 & $+13.89$ & $+5.56$ & 52.08 & $+28.12$ & $-38.54$\\
Healthy & 93.06 & 0.00 & $-6.94$ & 76.04 & 0.00 & 0.00\\
Combined$^\ast$ & -- & -- & -- & 55.47 & $+19.27$ & $-6.51$\\
\bottomrule
\end{tabular}
\caption{Frozen programs on unseen checkpoints, with no feedback. P1 and P3 were written from unrounded weights, P2 and P4 from rounded weights. Inputs are unrounded. LFM2-VL rows use L2; $^\ast$the four fault draws of L3 and L4. \emph{Start} is spatial accuracy (\%); other entries are changes in points. T1--T3 have 24 test questions each, the LFM2-VL checkpoints 96. On rounded copies, P1 and P3 make no edit, and P2 and P4 edit the same weights.}
\label{tab:transfer}
\end{table}

\paragraph{Programs written from unrounded weights.}
P1 and P3, about 100 lines each, implement the repair of Sec.~\ref{sec:residue} without labels. They estimate the factor of each off-grid row as the sessions do, P1 only for factors between $\tfrac12$ and 2, and shrink (P1) or zero (P3) sparse off-grid entries.

On unrounded scaling faults both programs change only damaged weights. P1 improves all three test checkpoints, by 16.67 to 25.00 points; it changes only weights that it restores exactly, which are 99.31\%, 98.61\%, and 100\% of the rescaled weights of T1, T2, and T3. P3 restores all 655,360 rescaled weights of L2 exactly, on the scaling fault and on the combined fault, and returns accuracy to the clean level or above (Tab.~\ref{tab:transfer}); on the four fault draws of L3 and L4 it restores at least 99.6\% of them and gains 19.27 points. Neither program makes an edit on healthy checkpoints, on sign-only faults, or on any rounded case.

\paragraph{Programs written from rounded weights.}
P2, from a session that did not locate the damage, rewrites visual position embeddings, which no fault touches, and lowers the accuracy of healthy checkpoints by 6.94 points. P4 rescales to the median every row of the LFM2-VL projector matrix whose RMS exceeds twice the median (Sec.~\ref{sec:hint}): on L2, 128 rescaled rows and no healthy one. It repairs the scaling fault on both copies without restoring a weight exactly and leaves the healthy checkpoint unedited. On the combined fault the same edit lowers accuracy from 52.08\% to 13.54\%, a loss of answer format: the model now continues after the answer word, and 23 of its 96 outputs are exactly \emph{yes} or \emph{no}, against 93 before. Counted by first word, a check we added after seeing these outputs, 52 answers are correct, against 51 before. On L3 and L4, P4 lowers accuracy by 6.51 points, or by 1.04 counted by first word, and the losses of the fixed rules of Sec.~\ref{sec:hint} are of the same kind. For every LFM2-VL result of a session or a fit, the two counts differ by at most four answers (Supp.~S22, S25).

\section{Discussion}
\label{sec:discussion}

\paragraph{What the agent uses.}
The agent's repairs depended on how the faults had been computed: with low bits that only damaged weights carry, it located the damage and usually restored the rescaled weights exactly, and without them it repaired neither model unaided. Its edits used four kinds of numerical evidence: the low bits; a distributional anomaly (empty bins in the fraction histograms of rescaled rows, and unusual row norms); structural redundancy (similar column blocks in the SmolVLM projector); and statistics across layers (a row's covariance with neighboring layers). None of the four requires knowing what the weights compute. The agents found the low bits unprompted on both models, and these alone revealed the sparse noise. They used the distributional anomaly unprompted where the scale factors were large, in two LFM2-VL sessions to the point of exact restoration, and on SmolVLM after a hint or misplaced low bits pointed to it. They used the last two only when told the kind of fault or shown its location. Where a session found none of these, its gain came from edits to healthy weights that shift the model's answers.

\paragraph{Limitations.}
We study two models, one agent model, synthetic scenes, and one family of questions about photographs. The checkpoints of a model differ only in a small adapted block and the planted faults. Apart from the combined fault on L3 and L4, every LFM2-VL condition has one session on one checkpoint, and differences of a few answers between such conditions are within what random control edits produce; on rounded L1 without marks, no agent edit differs from no edit by an exact McNemar test (Supp.~S27). L3 and L4 reuse the scale factors and the photographs of the earlier checkpoints, and one of their four rounded cells is empty. LFM2-VL answers are \emph{yes} or \emph{no}, so accuracy moves with the answer bias. The two testbeds differ in scale factors and budgets as well as in model and images, so a difference between them cannot be assigned to one of these. We wrote the fraction hint after inspecting A--C, and the grid-fit scripts, the histogram rule, and the gradient fits on T4--T9 are post hoc comparisons. We withhold the public checkpoint; an editor that held it and knew the adapted block could copy every other weight, which returns T1--T3 to clean accuracy on the combined fault (Supp.~S18). The fraction histogram depends on a multiplier shared across a row: per-weight factors or a rank-one update of the same size leave it at chance on the public SmolVLM projector. Every audit here depends on a clean reference, which real failures lack.

\paragraph{Implications.}
Perturbing a low-precision checkpoint in float32 arithmetic can leave nonzero low bits on the changed weights, as our scaling and noise faults do. We recommend that studies of weight repair report the number format of the weights and include a rounded control. Planted faults do not show whether natural failures leave such traces; a bit flip in a checkpoint stored in BF16~\cite{rakin2019bitflip,hong2019terminal}, for one, stays on the grid, as our rounded faults do.

\section*{LLM Usage Statement}

The author conceived the study, defined its research questions and scope, directed the work, and is responsible for its content and conclusions. LLMs assisted with implementation, execution, literature search, and editing.

{\small\bibliographystyle{ieeenat_fullname}\bibliography{references}}
\end{document}